\pdfoutput=1

\documentclass[11pt]{article}

\usepackage[final]{acl}

\usepackage{times}
\usepackage{latexsym}

\usepackage[T1]{fontenc}

\usepackage[utf8]{inputenc}

\usepackage{microtype}

\usepackage{inconsolata}

\usepackage{graphicx}

\RequirePackage{fancyhdr}
\RequirePackage{xcolor} 
\RequirePackage{algorithm}
\RequirePackage{algorithmic}
\RequirePackage{natbib}
\RequirePackage{eso-pic} 
\RequirePackage{forloop}
\RequirePackage{url}

\usepackage{xspace}
\usepackage{graphicx}
\usepackage{amsmath}
\usepackage{booktabs}
\usepackage{amsfonts}
\usepackage{amssymb}

\title{TrimLLM: Progressive Layer Dropping for Domain-Specific LLMs}

\author{Lanxiang Hu, Tajana Rosing, Hao Zhang \\
  University of California, San Diego \\
  \texttt{\{lah003, tajana, haozhang\}@ucsd.edu}
  }

\usepackage{multirow}

\newcommand{\sys}[0]{\textsc{TrimLLM}\xspace}
\newcommand{\vx}{\boldsymbol{x}}

\begin{document}

\maketitle

\begin{abstract}
Specializing large language models (LLMs) for local deployment in domain-specific use cases is necessary for strong performance while meeting latency and privacy constraints. 
However, conventional task-specific adaptation approaches do not show simultaneous memory saving and inference speedup at deployment time. Practical compression techniques like quantization and pruning require dedicated hardware or kernel support to achieve measured inference speedup. 
We develop \sys based on the \emph{layer-wise specialization} phenomenon we empirically observed and verified on contemporary LLMs. \sys reduces the depth of LLMs via progressive layer dropping. We show it retains LLMs' capacity in specific domains and achieves inference speedup irrespective of hardware and deep learning frameworks.
We evaluated \sys on LLMs of various sizes for inference; models adapted on medical, legal, and financial datasets all demonstrate $2.1-5.7\times$ inference speedup on consumer GPUs and up to $3.1\times$ speedup on A100 when compared to state-of-the-art model compression algorithms, with no loss in accuracy at 50$\sim$60\% model compression ratio.



\end{abstract}

\section{Introduction}
Large language models (LLMs) are increasingly prominent, evolving to serve specialized domains such as medicine~\citep{thirunavukarasu2023large}, law~\citep{yue2023disc}, and finance~\citep{wu2023bloomberggpt}. Their deployment in local environments is particularly valuable, addressing latency and privacy concerns, especially where sensitive data are involved.
For example, understaffed clinics greatly benefit from medical-specialized LLM assistants deployed locally. However, the substantial memory and computation required for inference present significant barriers to deploying specialized LLMs in such resource-limited scenarios. 


Post-training quantization (PTQ) has emerged as a key technique for adapting LLMs to resource-limited environments, by reducing weight bit precision to 4 or even 3 bits, with minimal degradation in model performance. 
However, the practical implementations of PTQ methods~\citep{dettmers2022llm, xiao2023smoothquant, frantar2022gptq, lin2023awq} depend on the availability of efficient kernels and vendor-specific hardware support for quantized computational operations. Unfortunately, such support is not widely accessible. In reality, applying many existing PTQ techniques oftentimes \emph{slows down} model inference on consumer-level hardware, as shown in Table~\ref{tab:benchmarking_on_hardware}. 
Similar results are seen with many pruning algorithms~\cite{kwon2022fast, frantar2023massive, sun2023simple}, which fail to translate theoretical speedup into real performance gains when specific hardware or kernel support (e.g., for sparsity) are absent.

\begin{figure*}[t]
    \center
        \includegraphics[width=0.49\linewidth]{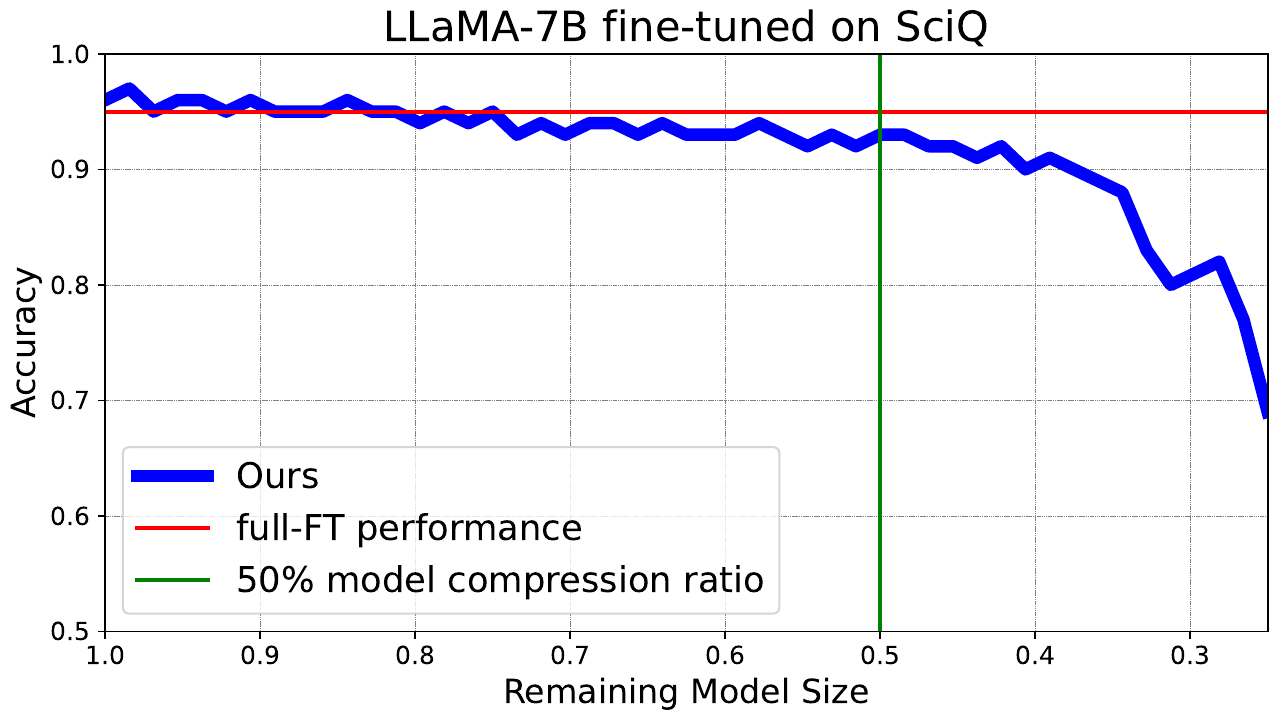}
        \includegraphics[width=0.49\linewidth]{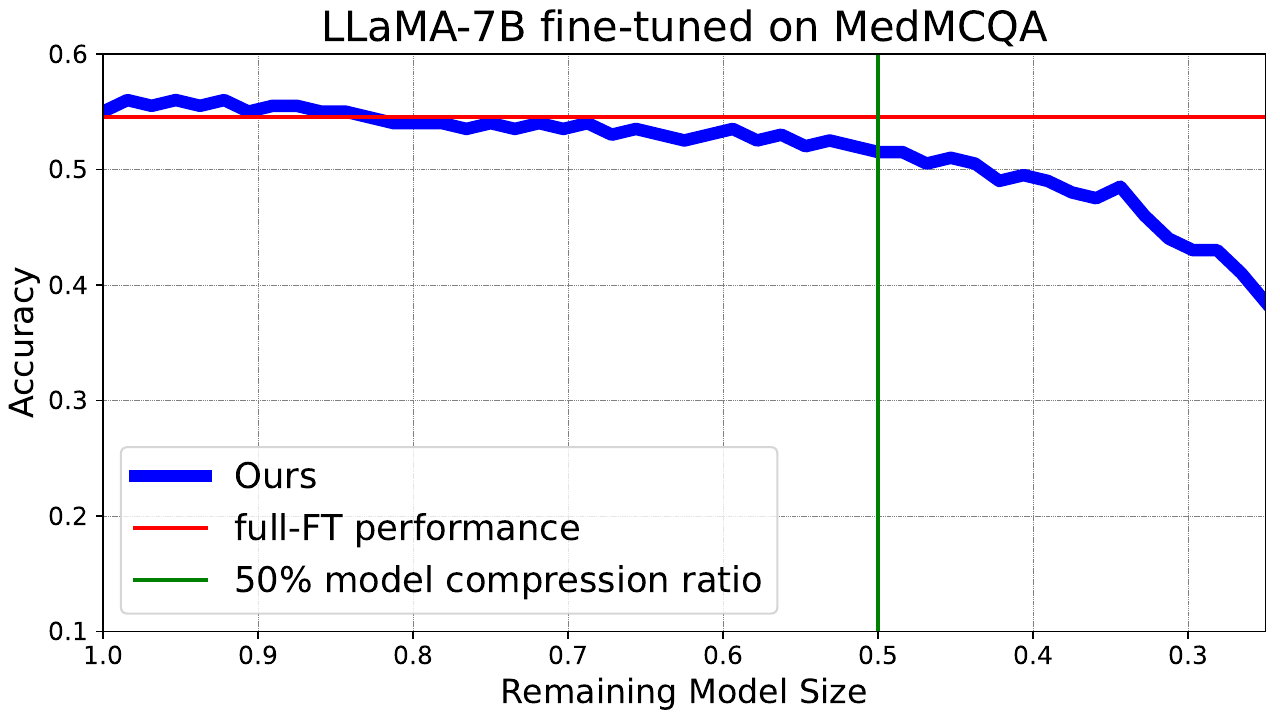}
        \caption{On SciQ and MedMCQA, LLaMA-7B can be reduced to $40\%\sim 50\%$ of its original size with nearly no loss in accuracy. The layer dropping strategy employed is with calibration scanning, activation-norm tie breaker, and sparse udpate at $r=\frac{1}{4}$.
        } 
       \label{fig:layer_dropping_teaser}
\vspace{-1.5\baselineskip}
\end{figure*}

To address these limitations, this paper explores a new way of compressing LLMs. Recent insights in LLM model editing show that middle layers in LLMs are crucial for domain-specific knowledge~\citep{meng2022locating, li2023pmet, azaria2023internal}, with attention modules handling general semantic correlations while MLP layers being more task-specific~\citep{geva2020transformer}.
In this study, we delve deeper into the domain-specific relevance of various layers in LLMs. Figure~\ref{fig:layer_dropping_teaser} reveals that when fine-tuning on science-common-sense and medical domains, we can remove up to 20 and 16 out of 32 layers respectively in LLaMA-7B without compromising performance.


Building on these findings, we hypothesize \emph{layer-wise specialization}: the significance of each layer of an LLM, particularly the MLP layer, varies according to the specific knowledge domain; we can fine-tune a more domain-focused LLM by selectively dropping layers unimportant to the targeted domain. This strategy enables us to craft models that are not only more compact but also finely balanced in terms of memory usage, inference speed, and domain-specific accuracy.

To validate this hypothesis, we conducted extensive layer-dropping experiments on domain-specific datasets \citep{pal2022medmcqa, chalkidis2021lexglue, FIQA}, where one least-important layer is removed after one epoch of fine-tuning. The results indicate that up to a significant number of the layers could be dropped during fine-tuning with negligible accuracy loss using an effective target selection algorithm.  Building on these findings, we introduce \emph{\sys}, a novel framework that combines fine-tuning with progressive layer-dropping. This approach employs a calibration dataset and an activation-based metric to efficiently identify and eliminate the most non-essential layers after each fine-tuning iteration. Remarkably, \sys can compress popular LLMs to less than 50\% of their original sizes while maintaining domain-specific performance on par with fully fine-tuned models, despite a drastic reduction in parameters. This results in a significant reduction in model depth, therefore memory and computational cost at inference.  Moreover, unlike PTQ or existing pruning methods, \sys does not introduce precision changes or sparse computation and needs no hardware support for measured speedup.


The key contributions of this paper are: 
\begin{list}{$\bullet$}{\leftmargin=1em \itemindent=0em}
    \item We observe and empirically validate the \emph{layer-wise specialization} phenomenon in contemporary LLMs. 
    \item We design \sys, a new model compression approach. \sys develops a new metric for quantifying layer importance and an algorithm to identify and eliminate layers of minimal importance during the fine-tuning process, compressing LLMs to less than $50\%$ of its original size without compromising its effectiveness. Our proposed method is orthogonal to the other model compression techniques. They can be used in combination with each other to achieve as much as $8\times$ model compression ratio as detailed in Section~\ref{sec:exp:combined_techniques}.
    \item We demonstrate \sys exceeds the efficiency of full-sized models in domain-specific applications, including medical, legal, and financial fields. We also show \sys's ability to realize $2.1-5.7\times$ inference speedup than the baseline quantization and pruning approaches on consumer-level hardware due to reduced model depth. An additional advantage of \sys is its ability to provide a flexible continuum of target model sizes, offering greater hardware adaptability than traditional compression methods.
\end{list}



\begin{table}[t]
\centering
\caption{
Deployment-time model inference overhead breakdown for LLaMA-7B on different GPUs, with sequence length 512 and batch size 1. The ``Mem'' entry refers to the ratio of final compressed model size versus the original model size in memory. \textbf{\sys consistently achieves a high inference throughput in all GPU types we test on}.
}
\vspace{0.5\baselineskip}
\small
\label{tab:benchmarking_on_hardware}
\resizebox{\linewidth}{!}{%
\begin{tabular}{c|c|c|c}
\toprule
 \text{Hardware}  & \text{Techniques}   & \text{Throughput (tokens/s)} & \text{Mem} \\ 
\midrule
\multirow{5}{*}{A100} & \text{FP16}   & 42.3 &  $100\%$ \\ 
& \text{SparseGPT}  & 58.9 & $100\%$ \\
& \text{LLM.int8()}   & 29.6 & $\geq50\%$ \\ 
& \text{GPTQ-int4}   & 46.5 & $\geq25\%$ \\ 
& \text{AWQ-int4}   & \textbf{115.3} & $\geq25\%$ \\ 
& \text{\sys}  &  \textbf{103.1}  & $\geq 40\%$ \\
\midrule
\multirow{5}{*}{V100} & \text{FP16}  & 16.6 &   $100\%$ \\ 
& \text{SparseGPT}  & 14.5 & $100\%$ \\
& \text{LLM.int8()}   & 10.2 & $\geq50\%$ \\ 
& \text{GPTQ-int4} & 6.1 & $\geq25\%$  \\ 
& \text{AWQ-int4} & 11.0 & $\geq25\%$ \\ 
& \text{\sys} & \textbf{34.9} &  $\geq 40\%$ \\  
\midrule
\multirow{5}{*}{RTX 3090} & \text{FP16}   & 13.4 & $100\%$ \\ 
& \text{SparseGPT}  & 13.0 & $100\%$ \\
& \text{LLM.int8()}     & 7.5  &  $\geq50\%$ \\ 
& \text{GPTQ-int4}   & 6.9  & $\geq25\%$ \\ 
& \text{AWQ-int4}   & 7.9  & $\geq25\%$ \\ 
& \text{\sys}  &  \textbf{26.8} &  $\geq 40\%$ \\ 
\bottomrule
\end{tabular}
}
\end{table}

\section{Related Work}

\noindent \textbf{Task-specific adaptation.}
A typical workflow for task-specific adaptation is to first fine-tune~\citep{wu2023pmc, yang2023fingpt, lawyer-llama-report, Lawyer-LLama} or even pre-train~\citep{wu2023bloomberggpt, cui2023chatlaw, shah2023creation} LLMs on task-specific datasets before applying any of the following three model compression techniques for reliable performance during inference: quantization, distillation, and pruning. In our case, we adopt layer-dropping to compress the model step-by-step \emph{during} fine-tuning, i.e., we adapt LLMs to domain-specific tasks by identifying and retaining important layers for the target domain.

\textbf{Quantization.} Quantization can effectively mitigate memory consumption by reducing the bit-widths of LLMs' weights and activations. Quantization has featured its ability to retain LLM’s zero-shot ability with measured memory saving and theoretical speedup. The state-of-the-art quantization algorithms~\citep{dettmers2022llm, xiao2023smoothquant} require implementations of efficient kernels whose efficiency relies on hardware support. To realize measured speedup for inference, decoding implementation for the specific quantization format is required ~\citep{dettmers2023spqr, lin2023awq}. \sys, on the other hand, does not depend on specialized kernels and it's making the model more efficient by reducing its depth. The performance gain can therefore be generalized to any hardware. 


\textbf{Pruning} Pruning aims to remove unimportant weights to reduce FLOPs. Latest post-training pruning algorithms for LLMs focus on structured and unstructured sparsity at neuron- or attention-head level~\citep{liu2023deja, sun2023simple, frantar2023sparsegpt, ma2023llm, ashkboos2024slicegpt} that need efficient kernels and hardware support for the corresponding structured sparsity patterns, without which it's hard to achieve measured efficiency improvement. \sys again requires none.

\textbf{Layer-dropping}. Layer-dropping is a relatively new technique in the context of LLM model compression. Some prior work investigate the feasibility of layer dropping by compressing a foundation model \emph{before it is fine-tuned} on downstream data~\citep{sajjad2023effect} or \emph{during the per-training stage}~\citep{zhang2020accelerating} (accelerate training with layer-dropping) to improve its efficiency. \sys conducts layer-dropping \emph{during fine-tuning}, reducing model size and adapting the model for specialized task simultaneously.

\noindent \textbf{Model Editing and Knowledge localization.}
\label{sec:related:knowledge_loc}
At layer-wise granularity, evidences~\citep{meng2022mass, frantar2023massive} show middle decoder blocks in LLMs contribute more to the domain-knowledge generation process while initial blocks are for low-level information (shallow patterns) extraction and last few blocks capture semantic patterns for next-token generation~\citep{azaria2023internal}. Within each decoder block, experiments ~\citep{geva2020transformer, meng2022locating} show that MLP layers are most responsible for task-specific memory retrieval and factual association. The attention layers, on the other hand, are meant to capture semantic correlation among all input tokens and therefore less specialized~\citep{shaw2018self}. \sys leverages different roles MLP and self-attention layers play to localize and drop the most insignificant layer.


\newtheorem{hyp}{Hypothesis}
\newtheorem{prop}{Proposition}
\section{Method}


\begin{algorithm}[tb]
\caption{\sys} 
\label{algo:sapling}
\small
\begin{algorithmic}[1]
\STATE {\bfseries Input:} 
Training data $\vx \in \mathcal X$ for the domain-specific task, pre-trained LLM $f(\cdot)$ with parameters $\theta$, training function $\mathcal F \left( \cdot \right)$ that optimizes some objective $\ell$, importance score metric $s$, sparse update ratio $r$, accuracy thresholding function $\mathcal C_a \left( a_i \right)$ or efficiency thresholding function $\mathcal C_e \left( M_i, T_i \right)$, $a_i$, $M_i$ and $T_i$ are model's accuracy, memory consumption and latency after the $i$-th layer is dropped. Buffers for sets $\mathcal A_{\mathcal X}$ and $\mathcal M_{\mathcal X}$ in Hypothesis~\ref{hypothesis}.
\STATE{$i \leftarrow 0$, $\mathcal A_{\mathcal X} \leftarrow \emptyset$, $\mathcal M_{\mathcal X} \leftarrow \emptyset$, $\mathcal U_{\mathcal X} := \mathcal A_{\mathcal X} \bigcup \mathcal M_{\mathcal X}$, $\theta_0 \leftarrow \theta$}
\STATE {$\mathcal{G}_{\mathcal U_{\mathcal{X}_0}} = f\left( \cdot \right)$, $n \leftarrow \text{total number of layers in } f(\cdot)$}
\STATE {\textbf{Sparse update:} Calculate initial $s_i$ for each layer. Freeze layers in accordance with $r$.}
\STATE {choose thresholding function $C \left( \cdot \right) \in \left\{ C_a, C_e \right\}$ that decides whether to exit}
\WHILE{not $C \left( \cdot \right)$}
    \STATE{Run training function to update the set of all parameters $\mathcal F \left( \cdot \right): \theta_i \to \theta_i'$}
    \STATE {$m \leftarrow 0$, $U \leftarrow \emptyset$}
    \WHILE{$m \neq n$}
        \STATE{Calculate layer-wise importance score $s_m$, append $s_m$ to $U$}
        \STATE{$m += 1$}
    \ENDWHILE
    \STATE {Choose which layer to drop with index $m$ s.t. $ s_m = \min(U)$, append $s_m$ to $\mathcal U_{\mathcal X}$}
    \STATE {Remove parameters: $\theta_i' \to \theta'_{i+1}$}
    \STATE {Remove layer $m$ an update the model: $\mathcal{G}_{\mathcal U_{\mathcal{X}_i}} \to \mathcal{G}_{\mathcal U_{\mathcal{X}_{i+1}}}$}
\ENDWHILE
\STATE {\textbf{Return: }{$\mathcal{G}_{\mathcal U_{\mathcal{X}}}$}}
\end{algorithmic}
\end{algorithm}

\subsection{Preliminaries and Layer-Wise Specialization}
\label{sec:method:prelim}

Auto-regressive language models compose of many transformer decoders, where each decoder block is made of one multi-head attention (MHA) layer and one MLP. Many previous studies on model editing (see Section~\ref{sec:related:knowledge_loc}) show increasing evidences suggesting that different layers weight differently when it comes to domain-specific inference, that we call \emph{layer-wise specialization}. 


Formally, consider a pre-trained model $f\left( \textbf{x} ; \mathbf{\theta}  \right)$, where $\textbf{x} \in \mathbb{R}^{s}$ is an input sequence with sequence length $s$ and embedding dimension $n$, $\mathbf{\theta} \in \mathbb{R}^D$ is a parameter vector that parameterizes $f \left( \cdot \right)$ with a total parameter size of $D$. 


Consider layernorm to be part of the MHA and MLP layer along with residual connection with each layer indexed by $i \in \left\{1, \dots, N \right\}$, where $N$ is the total number of layers in a model. Let the input to each decoder layer $\textbf{DEC}_i$ be $\textbf{y}_{i-1}$ at the current generation step, the corresponding output at layer $i$ follows expression in Eq.~\ref{eq:mha_mlp}.

\begin{align}
\textbf{y}_{i} = \textbf{DEC}_i\left( \textbf{y}_{i-1} \right) := \textbf{MLP}_i \left( \textbf{MHA}_i \left( \textbf{y}_{i-1} \right) \right)
\label{eq:mha_mlp}
\end{align}

At $i=1$, the input has $\textbf{y}_{i-1} = \textbf{y}_0 = \left( y_{0,1}, \dots, y_{0,T} \right)$, where $T$ is the current timestamp and $y_t$ is token generated by a previous timestamp $t < T$. 

Let the feature space for inputs of a downstream task be $\mathcal{X}$ and input tokens $y_{0,t} \in \mathcal{X}$, and the feature space for generated output tokens be $y_{N,t} \in \mathcal{Y}$ in Equation~\ref{eq:final_output}.

\begin{equation}
\begin{split}
\textbf{y}_{N} &= \textbf{DEC}_N \circ \textbf{DEC}_{N-1} \circ \dots \circ \textbf{DEC}_0 \left( \textbf{y}_0 \right) \\
&= f \left( \textbf{y}_0 ; \mathbf{\theta} \right)
\end{split}
\label{eq:final_output}
\end{equation}

Our basic assumption is that for each downstream task, there exists a feature space $\mathcal{X}$, where $\mathcal{X}$ can be described as a random variable from a distribution $D_{\mathcal{X}}$, and $\mathcal{Y}$ is a random variable from $D_{\mathcal{Y}}$. Our hypothesis is:

\begin{hyp}
\label{hypothesis}
Let the set of all attention layers in Equation~\ref{eq:mha_mlp} be $\mathcal{A}$ and the set of all MLP layers be $\mathcal{M}$. For all input sequences $x_0$ generated from $\mathcal{X}$, there exists a set of attention and MLP layers $\mathcal{A}_{\mathcal{X}} \subset \mathcal{A}$, $\mathcal{M}_{\mathcal{X}} \subset \mathcal{M}$ such that the function composition of $U_{\mathcal{X}} = \mathcal{A}_{\mathcal{X}} \bigcup \mathcal{M}_{\mathcal{X}}$ can be fine-tuned on the joint distribution $D_{{\mathcal{X}}{\mathcal{Y}}}$ for the downstream task to get a function $\mathcal{G}_{U_{\mathcal{X}}}$ with $\mathcal{G}_{U_{\mathcal{X}}}(\textbf{y}_0) =  \textbf{y}’_{N}$. It suffices that output of the model $\textbf{y}’_{N}$ is generated with random variable $\mathcal Y'$ from $D_{\mathcal Y'}$ and $D_{\mathcal Y'}$ is a close approximation of $D_{\mathcal Y}$ for the full model.
\end{hyp}

 Note that the order of function composition for $U_{\mathcal{X}}$ is in accordance with their original order in Equation~\ref{eq:mha_mlp}. 



\subsection{Fine-Tuning with Progressive Layer Dropping}
\label{sec:method:pipeline}

In addition to the ordinary fine-tuning procedure for language models, \sys iteratively picks a layer to drop after one epoch of training and gradually reduces the model depth. This gives \sys the advantages of reduced memory consumption and inference latency at deployment time. 

Our empirical experiments and recent works~\citep{syed2023prune} show drastically changing the model from $f(\textbf{y}_0; \theta_0) \to \mathcal{G}_{\mathcal U_{\mathcal{X}}}(\textbf{y}_0; \theta_f)$ by dropping many parameters all at a time generally gives bad results. This function $\mathcal{G}_{\mathcal U_{\mathcal{X}}}(\textbf{y}_0; \theta_f)$ maps the generated outputs to a distribution $D_{{\mathcal Y}_f}$ that’s very distinct from $D_{\mathcal Y}$ and result in bad domain-specific performance. Note that $\theta_f$ is the parameter vector and $D_{\mathcal Y}$ is the output distribution for the full model after fine-tuning. Successive layer dropping, on the other hand, allows domain-specific specialization to be done step by step with $f(\textbf{y}_0; \theta_0) \to \mathcal{G}_{\mathcal U_{\mathcal{X}_1}}(\textbf{y}_0; \theta_1') \to \mathcal{G}_{\mathcal U_{\mathcal{X}_2}}(\textbf{y}_0; \theta_2') \dots \to \mathcal{G}_{\mathcal U_{\mathcal{X}}}(\textbf{y}_0; \theta_f')$ where $\theta_i'$ is the parameter vector after $i$ epochs. $\mathcal{G}_{\mathcal U_{\mathcal{X}_i}}(\cdot)$ is the model right after the $i$-th epoch with the corresponding set of remaining layers being $U_{\mathcal{X}_i}$. 

This observation aligns the intuition that gradually changing the function's parameterization with most important layers retained allows generated outputs to transit more smoothly from $D_{{\mathcal Y}_0}' \to D_{{\mathcal Y}_1}' \to \dots \to D_{{\mathcal Y}_f}'$ such that $D_{{\mathcal Y}_f}'$ is a close approximation of $D_{{\mathcal Y}}$ for the full model after fine-tuning. It thereby provides more evidences to verify our hypothesis in Section~\ref{sec:method:prelim} with an additional constraint: 

\begin{prop}
\label{successive_distribution_shift}
The functional $\mathcal R: f(\cdot) \to \mathcal{G}_{U_{\mathcal{X}_i}}(\cdot)$ needs to be decomposed into successive layer-dropping operators $\left\{ r_0, \dots, r_f \right\}$ such that the parameter vector $\theta_i'$'s dimensionality only changes by a small decrement at a time to gradually adapts a downstream task with the most representative parameters. 
\end{prop}


Due to the iterative nature of the aforementioned layer dropping algorithm, the time complexity of fine-tuning increases as more layers are to be dropped. We address this training cost and reduce it to the same order of magnitude as baseline FT that trains for only a few epochs. This is accomplished by using two techniques: sparse fine-tuning as delineated in Section~\ref{sec:method:sparse_update} and adaptive layer dropping. See Section~\ref{appendix:time-complexity} further details. In practice, this approach enables users to efficiently exchange a slightly longer model adaptation time for improved inference-time performance, which aligns with the typical development-deployment cycle observed in many real-world applications. In such cases, developers often have the flexibility to accommodate longer development periods but place higher demands on deployment-time performance. 

\begin{figure*}[t]
    \centering
        \includegraphics[width=0.49\linewidth]{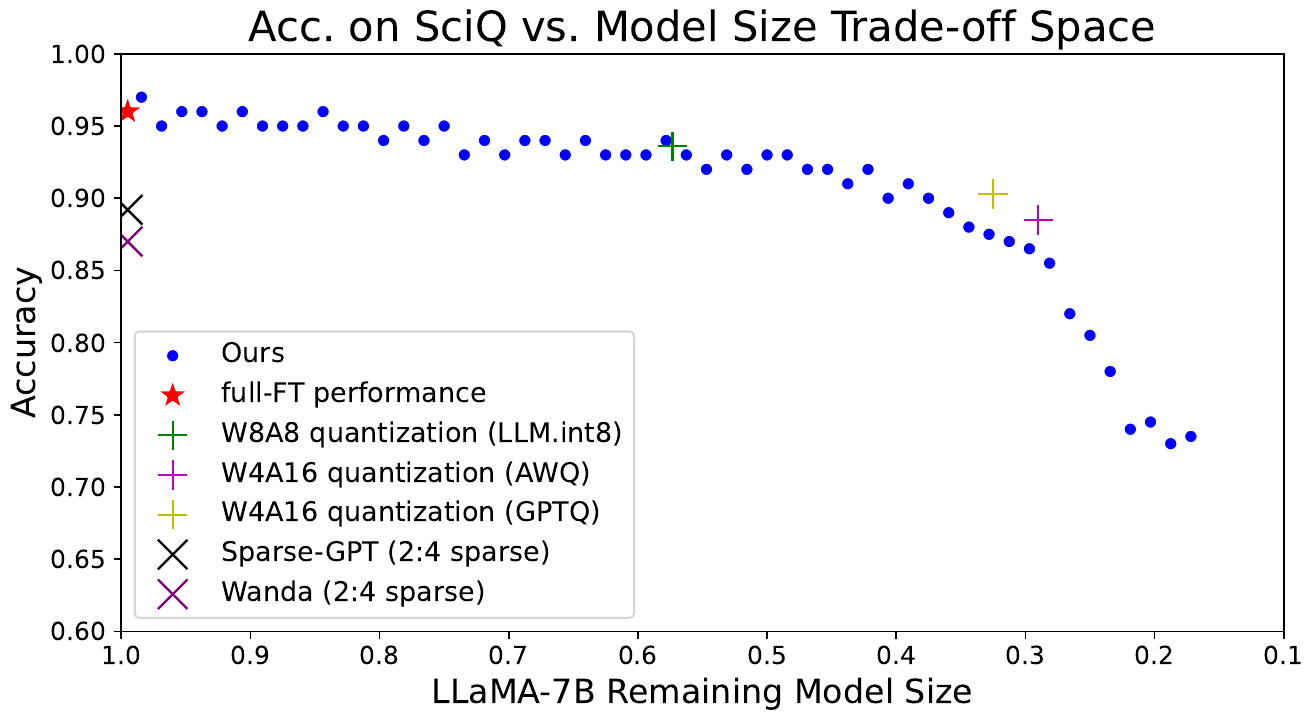}
        \includegraphics[width=0.49\linewidth]{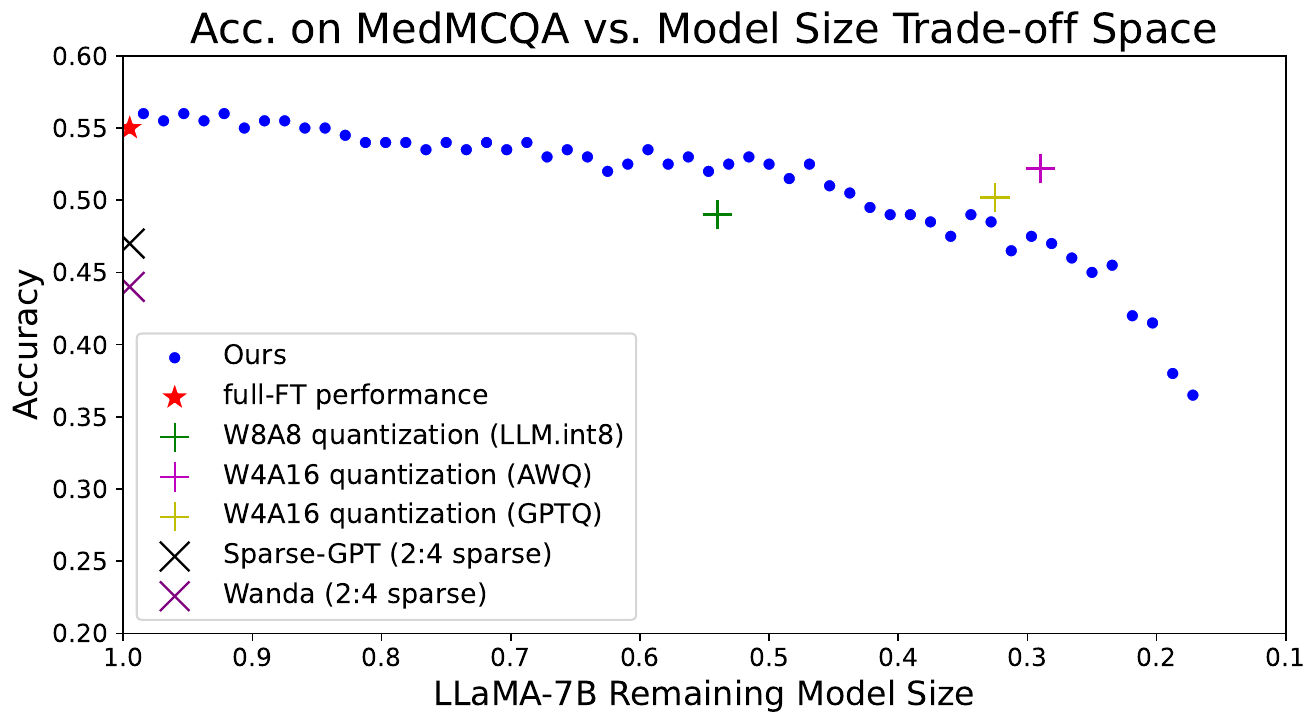}
    \caption{The Pareto Frontier of LLaMA-7B-\sys on SciQ and MedMCQA. \sys has a much wider spectrum of operating points to fit the model into different hardware with competitive performance. The layer dropping strategy employed is with calibration scanning and activation-norm tie breaker and + sparse udpate at $r=\frac{1}{4}$.}
    \label{fig:operating_points}
    \vspace{-1\baselineskip}
\end{figure*}

\subsection{Target Selection Algorithms}
\label{sec:method:target_selection}

One important aspect of \sys is choosing the right layer from $U_{\mathcal{X}_i}$ to drop after the $i$-th epoch and thereby satisfy the successive distribution shift condition (Proposition~\ref{successive_distribution_shift}). We introduce two techniques to assign each layer an importance score, where a lower importance score means the layer contribute less to the model's performance on a downstream task.

\textbf{Sensitivity-based Scoring}. The first method is a performance scanning based on a small calibration dataset. Before each time a layer is to be dropped, a small subset of the fine-tuning dataset's validation set is sampled as the calibration dataset. For each layer, its importance score is the reciprocal of the model's performance after dropping the layer. Calibration scanning gives the importance score of any layer $i$ and the expression is presented in Equation~\ref{eq:scan_score}, where $a_i \in [0, 100]$ is the accuracy of the model after dropping the $i$-th layer and $\delta$ is a small positive number such that $\frac{100}{1 + \delta^2}$ is the maximum importance score when $a_i = 0$.

\begin{equation}
s_{i, \text{scan}} = \dfrac{100-a_i}{\left( 1+\delta^2 \right) + \left( 1+\delta \right)a_i}
\label{eq:scan_score}
\end{equation}

\textbf{Activation-based Scoring}. The second method is to make activation-norm comparison on different layers' activations. Recent studies~\citep{dettmers2022llm, xiao2023smoothquant} have shown preserving information carried by activations is critical to model's performance when it comes to model compression techniques. 

In our work, our goal is to only preserve activations that are meaningful to the knowledge domain of interest. We can drop the rest to trade the model's generality for efficiency and specialization. A new metric is therefore needed to quantify the importance of an activation. 

Our assumptions consist of two parts: (1) there exists a feature space $\mathcal X$ and a corresponding low intrinsic dimension~\citep{aghajanyan2020intrinsic}. (2) activation tensors are dense with mostly small-magnitude elements and a few large-magnitude outliers based on widely recognized observations~\citep{dettmers2022llm, xiao2023smoothquant}.

Among common matrix norms including the $\ell_{2,1}$ norm, the Forbenius norm and the nuclear norm, at the same numerical value, nuclear norm should be the best metrics for directly measuring the rank of a matrix which is defined as the sum of the singular values of the matrix: $||W||_* = \sum_i \sigma_i$. The nuclear norm is a convex surrogate for the rank function and is often used in rank minimization problems. However, The nuclear norm introduces extra computational overhead because it requires the computation of the SVD of the matrix. Computing the SVD is computationally intensive, especially for large matrices, as it has a complexity of $O(\text{min} (nm^2,mn^2))$ for $m\times n$ matrix. As a result, we use the Forbenius norm to approximate the nuclear norm. By expanding the Forbenius norm with SVD, it follows: $||W||_F = \sqrt{ \sum_i \sigma_i^2 }$.

Therefore, we choose the Forbenius norm to identify activations with high-rank representations and sparse domain-specific knowledge. Dropping the one with highest norm is analogous to Forbenius norm minimization. Let $\left\{ {\lVert \vx_j \rVert_F} \right\}$ be the set of Forbenius norm for all remaining layers in the model $f \left( \cdot \right)$. This activation-norm importance score can be expressed in the form of Equation~\ref{eq:norm_score} such that $s_{i,\text{norm}} \in (0, 100]$. 

\begin{equation}
s_{i,\text{norm}} = \dfrac{100 \min \left\{ {\lVert \vx_j \rVert_F} \right\}}{\lVert \vx_i \rVert_F}
\label{eq:norm_score}
\end{equation}

\subsection{Sparse update as a Regularization}
\label{sec:method:sparse_update}

In \sys, an important observation is that some less important layers will eventually be dropped regardless whether they have been tuned.  Moreover, empirical evidences in Table~\ref{tab:method_compare_performance_llama} show fine-tuning all layers could, in effect, perform worse than full fine-tuning.

There are two reasons for the possible performance degradation. First, catastrophic forgetting has been a well recognized problem when a language model is trained on downstream data with all parameters are updated~\citep{lee2022surgical}. 
Second, layer dropping in \sys is conducted on the premise that some layers carry less information for a task and can be discarded. However, fine-tuning all layers is based on a contradictory premise that all layers need to be updated for downstream adaptation. As a result, it’s natural to adopt a sparse update scheme where we only update the layers with greatest chance to be kept after layer dropping. 

To identity which layers to be updated and which to be frozen, we run layer-wise importance score scanning with a calibration dataset before any fine-tuning is done. This gives an initial distribution of all layers' importance scores and probability to be dropped in the first epoch. According to Section~\ref{appendix:layer_dropping_patterns}, Layer-dropping patterns, since the initial distribution is highly correlated with the latter ones, we can assume fine-tuning with layer dropping won't significantly disturb each layer's importance score and use this initial distribution to infer each layer's overall probability to be dropped. For a sparse update ratio $r$, only up to $N'= r \times N$ layers will be updated in \sys. It's possible for any of the $N'$ layers to be dropped during fine-tuning. Each time this occurs, no additional layers will be made trainable.


\section{Experiments}

In this section, we present experiments that provide empirical evidences for our hypothesis as well as the effectiveness of \sys. The test suite spans a wide range of knowledge domains including common-sense, medical, legal and financial QA benchmarks. All experiments reported in this section are conducted on LLaMA-7B and LLaMA-13B with training performed on NVIDIA V100 32GB servers. Deployment-time inference speeds are tested on NVIDIA A100 40GB, V100 32GB and RTX 3090 GPUs.


\subsection{Performance on QA Benchmarks} 

To test which of the methods can compress the model to the fullest extent while maintaining more than $90\%$ performance of the full-finetuning baseline, we compare the performance of different sparse update schemes and target selection algorithms. The results are summarized in Table~\ref{tab:method_compare_performance_llama}. On each QA benchmark, we also compare \sys and other model compression techniques. The results are presented in Table~\ref{tab:qa_performance_llama}.


\begin{table*}[t]
\centering
\caption{
Performance comparison of LLaMA-7B and LLaMA-13B variants on QA benchmarks. The numerical values are percentage in accuracy. \sys here uses the best strategy with sparse update at $r = \frac{1}{4}$, calibration scanning and activation-norm tie breaker. For sparse-FT, the frozen layers are determined by calibration scanning and $r=\frac{1}{4}$. The ``Final Mem'' entry refers to the ratio of final compressed model size versus the original model size in memory. The ``LLM-Pruner'' baseline here uses the best-performing element-wise method at 50\% sparsity ratio. The ``SparseGPT'' baseline uses 2:4 structured sparsity at 50\% sparsity ratio, which yields the most significant latency reduction on most hardware.
}
\vspace{0.5\baselineskip}
\small
\label{tab:qa_performance_llama}
\begin{tabular}{ccccccc}
\toprule
\text{models}  & \text{PIQA} & \text{SciQ} & \text{MedMCQA}  & \text{LexGLUE} &  \text{FinanceQA}  & \text{Final Mem $\left( \downarrow \right)$}        \\ 
\midrule
\multicolumn{7}{c}{LLaMA-7B} \\
\midrule
\text{w/o training}  &  77.4 & 89.7 & 22.4 & 32.1 & 33.6 & 100\% \\ 
\text{+ Full-FT} & 82.4 & 95.6 & 54.6 & 42.9 & 45.1 & 100\% \\ 
\text{+ Sparse-FT} & 83.1 & 95.4 & 53.7 & 43.4 & 46.9 & 100\% \\ 
\midrule
\text{+ LLM-Pruner} & 70.3 & 85.0 & 23.1 & 30.8 & 27.3 & $100\% $ \\
\text{+ SparseGPT (2:4)} & 76.5 & 90.1 & 52.3 &  37.9 & 41.6 & $100\% $\\ 
\text{+ LLM.int8()} & 81.7 & 93.6 & 52.0 & 40.9 & \textbf{44.9} & $> 50\%$ \\
\text{+ AWQ-int4} & 80.9 & 93.0 & 50.7 & 41.0 & 42.1 & $> 25\%$ \\ 
+ \sys(50\%) & \textbf{81.8} & \textbf{94.2} &  \textbf{53.1} & \textbf{42.0} & 43.6 & $\geq 50\%$ \\ 
+ \sys (40\%) & 77.6 & 91.2 & 47.5 & 39.5 & 41.3 & $\geq 40\%$ \\ 
+ \sys (30\%) & 68.5 & 87.3 & 45.8 & 36.8 & 36.0 & $\geq 30\%$ \\
\midrule
\multicolumn{7}{c}{LLaMA-13B} \\
\midrule
\text{w/o training}  & 79.4  & 92.0 & 24.0 & 37.2 & 35.3 & 100\% \\ 
\text{+ Full-FT} & 83.9 & 97.2  & 57.2 & 48.3 & 50.1 & 100\% \\ 
\text{+ Sparse-FT} & 84.1 & 97.6 & 56.5 & 48.0 & 49.5 & 100\% \\ 
\midrule
\text{+ LLM-Pruner} & 65.3 & 80.2 & 45.2 & 33.0 & 29.5 & $100\% $\\
\text{+ SparseGPT} & 76.0 & 95.4 & 53.6 &  40.8  & 48.1  & $100\% $\\ 
\text{+ LLM.int8()} & 81.5 & \textbf{96.0} & 54.2 & 45.5 & 48.4 & $> 50\%$ \\
\text{+ AWQ-int4} & 80.5 & 95.9 & 53.5 & 44.1 & 45.6 & $> 25\%$ \\ 
+ \sys (50\%) & \textbf{82.4} & 95.8 & \textbf{56.9}  & \textbf{46.3}  & \textbf{47.5} & $\geq 50\%$ \\ 
+ \sys (40\%) & 79.1 & 93.5 & 50.1  & 43.8  & 45.8 & $\geq 40\%$ \\ 
+ \sys (30\%) & 76.2 & 91.2 & 47.4  & 39.7  & 41.0 & $\geq 30\%$ \\
\bottomrule
\end{tabular}
\vspace{-0.5\baselineskip}
\end{table*}

\begin{table*}[th]
\centering
\caption{
Performance comparison of LLaMA-7B \sys variants on QA benchmarks with various target selection algorithms (Section~\ref{sec:method:target_selection}). The optimal sparse update ratio at $r=\frac{1}{4}$ is used. For sparse-FT, the frozen layers are determined by calibration scanning and $r=\frac{1}{4}$. Check Table~\ref{tab:complete_method_compare_performance_llama} for a comprehension ablation on different sparse-update ratios.
}
\small
\label{tab:method_compare_performance_llama}
\begin{tabular}{ccccccc}
\toprule
\text{methods}  & \text{PIQA} & \text{SciQ} & \text{MedMCQA}  & \text{LexGLUE} &  \text{FinanceQA}  & \text{Final Mem} $\left( \downarrow \right) $       \\ 
\midrule
\multicolumn{7}{c}{ \text{LLaMA-7B} }\\
\text{w/o fine-tuning}  &  77.4 & 89.7 & 22.4 & 32.1 & 33.6 & 100\% \\
\text{+ Full-FT} & 82.4 & 95.6 & 54.6 & 42.9 & 45.1 & 100\% \\ 
\text{+ Sparse-FT} & 83.1 & 95.4 & 53.7 & 43.3 & 46.9 & 100\% \\ 
\midrule
\multicolumn{7}{c}{LLaMA-7B-\sys $\left(r=\frac{1}{4}\right)$}  \\ 
+ calibration  & 80.5 & 94.0 & 52.4 & 41.5 & 42.3 & $\geq 55\%$ \\ 
+ activation-norm  & 79.6& 93.5 & 51.5 & 39.8 & 41.7 & $\geq 85\%$ \\ 
+ both & \textbf{81.8} & \textbf{94.2} & \textbf{53.1} & \textbf{42.0} & \textbf{43.6} & $\mathbf{\geq 50\%}$ \\
\midrule
\multicolumn{7}{c}{LLaMA-7B Rule-based Layer Dropping \citep{sajjad2023effect}}  \\ 
+ random 50\% layers  & 65.4 & 78.5 & 35.0  & 21.3 & 21.3 & $\geq 50\%$ \\ 
+ top 50\% layers & 68.7 & 82.2 & 34.2 & 23.1 & 26.0 & $\geq 50 \%$ \\ 
+ bottom 50\% layers & 50.6 & 63.9 & 20.3 & 17.4 & 16.8 & $\geq 50 \%$ \\
\bottomrule
\end{tabular}
\end{table*}

\textbf{Baselines}. We use full fine-tuning (full-FT) as our most basic baseline. We also include a sparse fine-tuning (sparse-FT) baseline that only updates the salient layers identified by calibration scanning with the optimal sparse update ratio $\left( r = \frac{1}{4}\right)$. While LLM pruning approaches with structured pruning methods can give inference speedup as shown in Table~\ref{tab:benchmarking_on_hardware}, they are generally incapable of reducing memory consumption without hardware support. As a result, we benchmark \sys with the state-of-the-art LLM quantization techniques: LLM.int8(), GPTQ and AWQ. They are used as stronger baselines that permit both memory saving and potential inference speedup. 

\textbf{QA benchmarks}. We use common-sense QA benchmarks inculuding SciQ~\citep{SciQ} and PIQA~\citep{Bisk2020} to test LLM's ability of understanding and making basic inference about the physical world the way ordinary humans do. To further assess \sys's capacity for domain-specific adaptation, we also evaluate its performance on medical, legal, and financial QA datasets: MedMCQA~\citep{pal2022medmcqa}, LexGLUE-casehold~\citep{chalkidis2021lexglue}, and FinanceQA~\citep{alpaca-finance} respectively. For LexGLUE, evaluations are done on the "law" subset of MMLU~\citep{hendrycks2020measuring}. For FinanceQA, the dataset includes a combination of FiQA~\citep{FIQA}, Stanford-Alpaca~\citep{alpaca}, and ChatGPT QA dialogues. Evaluations of are conducted on the "economics" subset of MMLU for its pertinence to financial knowledge. Cross-evaluations are conducted on the PubMedQA~\citep{jin2019pubmedqa} and Legalbench~\citep{guha2023legalbench} benchmarks to test the specialized models' performance on a similar knowledge domain. All experiments adhere to established academic evaluation standards, utilizing the \texttt{lm-evaluation-hardness}\footnote{The repository is publically available at \url{https://github.com/EleutherAI/lm-evaluation-harness}.} repository.

\textbf{Results}. In addition to the two target selection methods introduced in Section~\ref{sec:method:target_selection}, we device a new two-step algorithm that leverages both methods, which corresponds to the entry “both” in Table~\ref{tab:method_compare_performance_llama}. This method adopts the more effective calibration scanning as the primary method for layer dropping target selection and uses activation-norm comparison as the tie-breaker strategy when there are more than one layer have the same importance score from calibration scanning. We can see from Table~\ref{tab:method_compare_performance_llama} the two-step algorithm gives the best specialized model at every sparse update ratios.


For each of the three methods, we evaluate specialized models performance when they are trained with different sparse update ratio $r = \left\{ 1, \frac{1}{2}, \frac{1}{4}, \frac{1}{8} \right\}$. As we can see in Table~\ref{tab:method_compare_performance_llama}, in comparison with other target selection techniques, we find at a sparse update ratio of $r = \frac{1}{4}$, the model performs the best. See Section~\ref{appendix:sparse_update} for detailed ablation on different values of $r$. The results also show that all three rule-based methods employed in \citet{sajjad2023effect} perform very poor in comparison with \sys. 

To demonstrate \sys's effectinveness on other LLMs, experiments are conducted on OPT-1.3B and OPT-6.7B for one common-sense, medical, and legal benchmark each, in Table~\ref{tab:qa_performance_opt}. The results validate the generalizability of our method as its effectiveness on the OPT models.


\subsection{Memory Consumption and Latency} 

We argue the \sys has a two-fold advantage. The first one is efficiency and the other is flexibility.

On the efficiency side, \sys has both deployment-time memory saving and inference speedup. We compare \sys with other model compression baselines as shown in Table~\ref{tab:benchmarking_on_hardware} and Figure~\ref{fig:operating_points}. The state-of-the-art quantization techniques are able to reduce inference-time memory consumption to nearly a quarter in size. \sys exploits the model depth degree of freedom and is able to achieve competitive memory saving compared to the quantization baselines with faster inference speed (Table~\ref{tab:benchmarking_on_hardware}) on consumer-level hardware, V100 and RTX 3090 GPUs, where hardware support for low-precision inference and structured sparsity are unavailable in the tensor cores.

On the flexibility side, as we can see from Figure~\ref{fig:operating_points}, quantization and pruning offers a very limited set of operating points corresponding to each of the bit precision scheme for each model. Since sparsity ratio in pruning can not be easily translated into memory saving, pruning oftentimes gives even fewer operating points in the trade-off space. In contrast, the Pareto frontiers of \sys span a wide range of operating points. As a result, \sys is more flexible and is capable of fitting a model to a wide spectrum of hardware. 


\subsection{Applying Other Model Compression Techniques to \sys}
\label{sec:exp:combined_techniques}

Our method is orthogonal to all model compression techniques. Applying \sys alongside with other post-training model compression techniques like quantization can provide further speedup. Results from applying AWQ and SparseGPT with 2:4 structured sparsity to \sys, and the corresponding compressed models' accuracy, memory consumption and inference latency are reported in Table~\ref{tab:composite_qa_performance_llama} with LLaMA-7B across a variety of benchmarks. 

The results show that on A100, other model compressible techniques can be seamless integrated with \sys to obtain even more efficient domain-specific models with as much as $8\times$ model compression ratio in terms of memory consumption, which translates to $4.5\times$ speedup in comparison with the uncompressed baseline.

\subsection{Limitations}
\label{sec:exp:limitations}

While increasing \sys can extend LLM accessibility to a wider audience in domain-specific use cases, specializing LLMs may raise robustness concern when applying the models to tasks that require knowledge from multiple domains. Striking a balance between accessibility and maintaining the integrity and reliability of language models is essential to ensure their responsible use in various applications.
\section{Conclusion}

We propose \sys, a task-specific adaption and model compression pipeline for contemporary LLMs. \sys reduces deployment-time memory cost and inference latency by identifying and discarding less significant layers to reduce the specialized model's depth. 
Unlike baselines, \sys can obtain both wall-clock inference speedup and memory saving without the need for specialized hardware and efficient computational kernels. We hope that \sys paves the path for making LLMs accessible to the wider public in personal and professional use cases.


\newpage
\section{Use of AI Assistants}

In adherence to the ACL Publication Ethics Policy, we did not employ AI assistants to generate the initial draft of this paper. We used AI assistants (GPT-4o) exclusively at the sentence level to enhance writing quality and correct grammatical errors.
\bibliography{TrimLLM}

\appendix
\appendix
\onecolumn
\section{Appendix}

\subsection{Layer-dropping Patterns} 
\label{appendix:layer_dropping_patterns}

\begin{figure*}[h]
    \centering
    \includegraphics[width=0.245\linewidth]{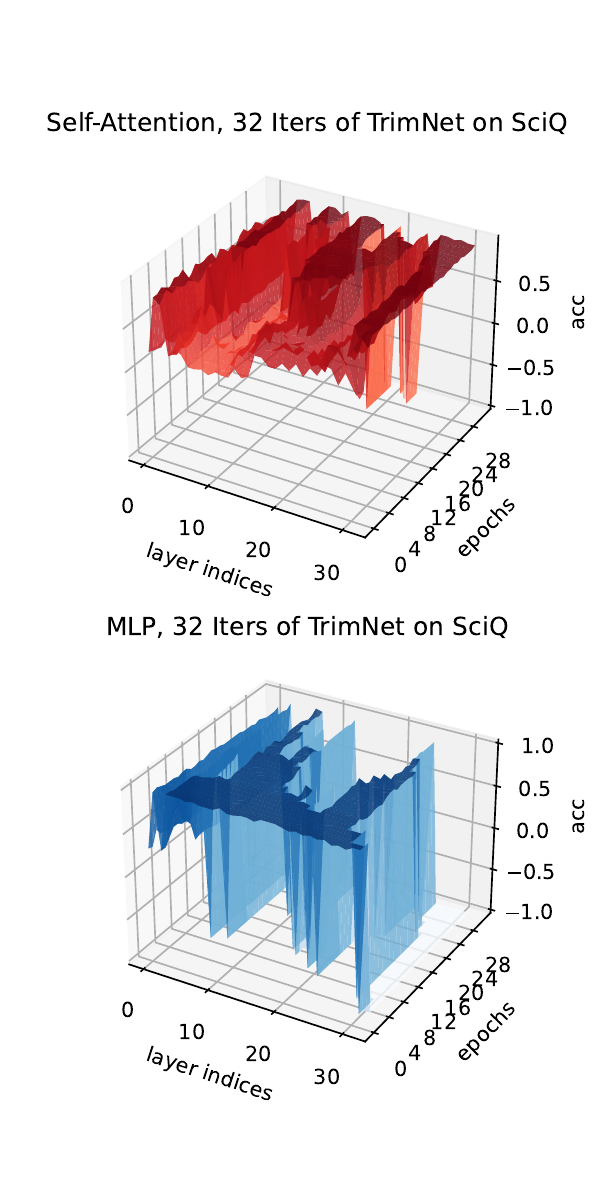}
    \includegraphics[width=0.245\linewidth]{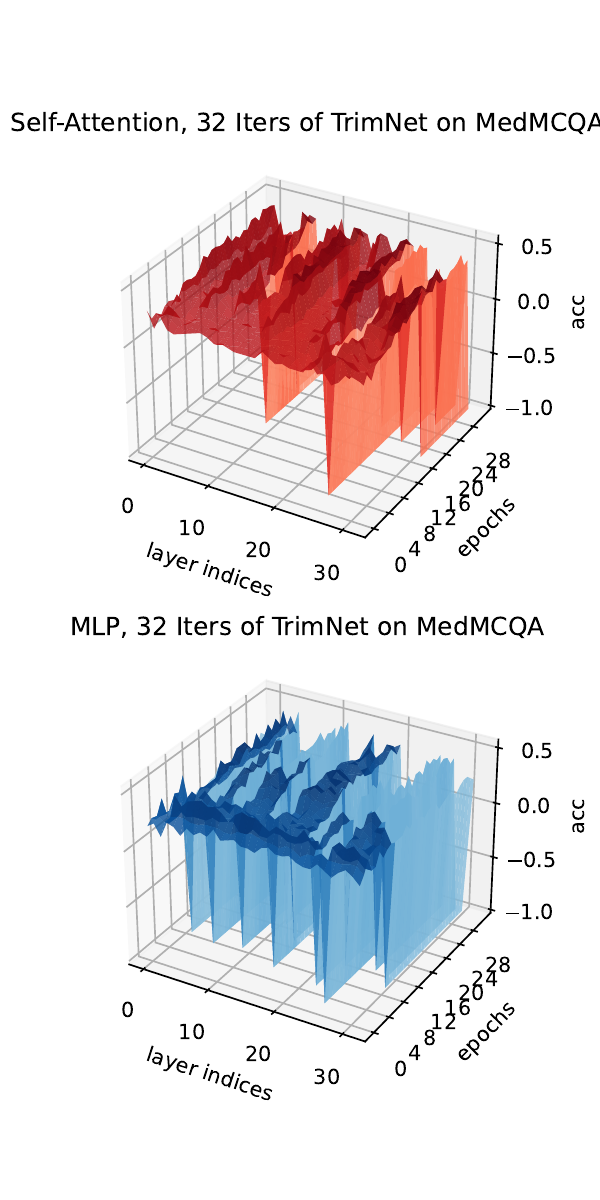}
    \includegraphics[width=0.245\linewidth]{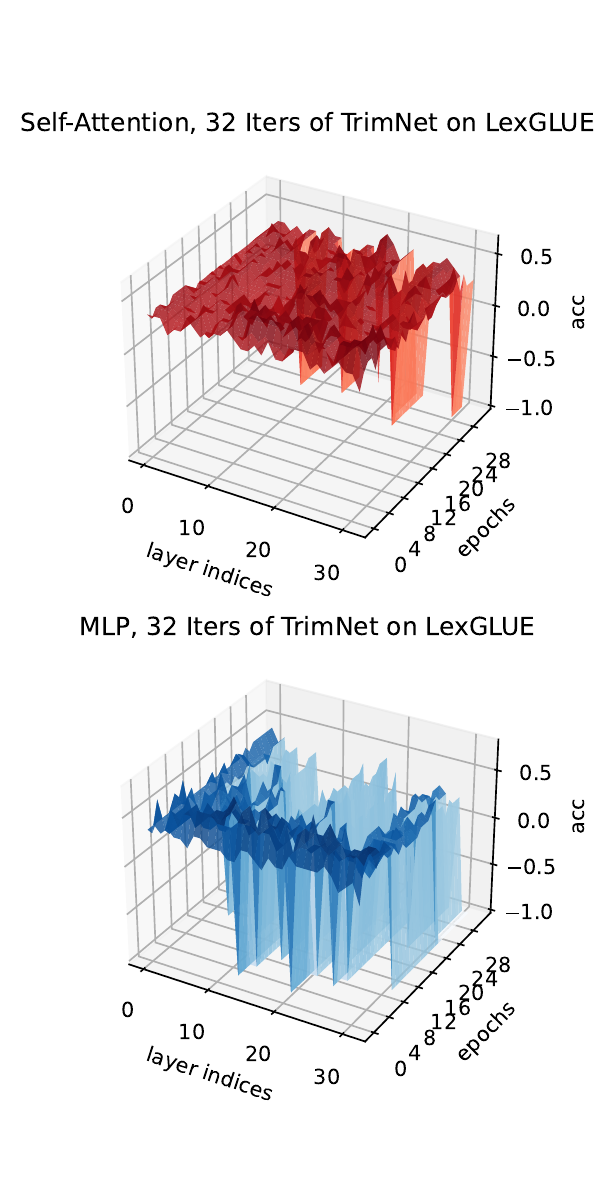}
    \includegraphics[width=0.245\linewidth]{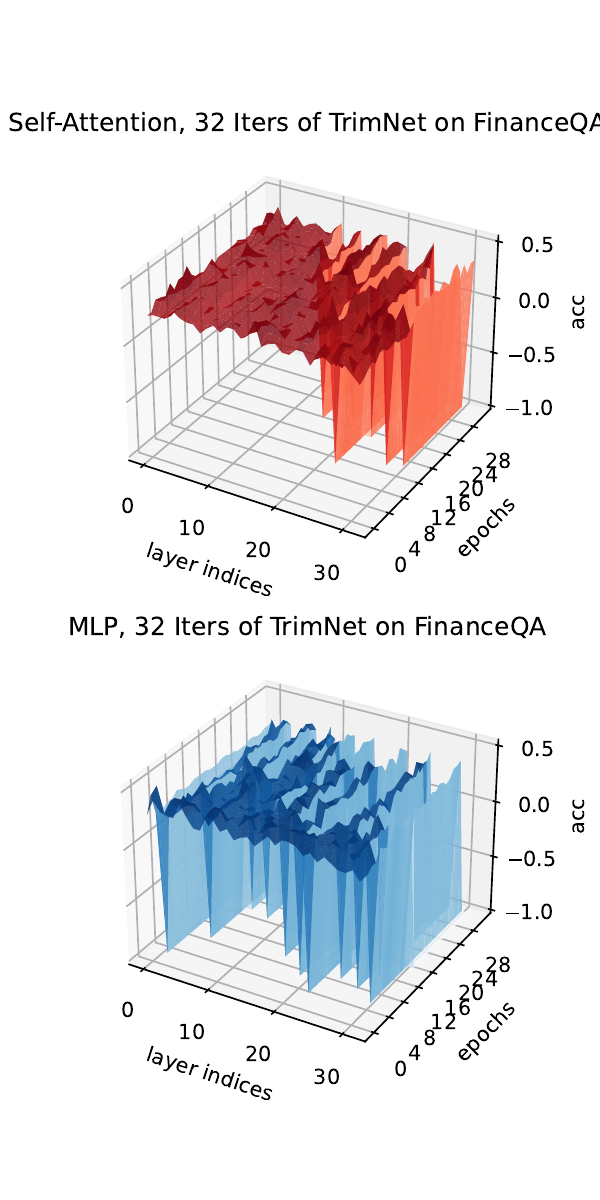}
    \vspace{-1.5\baselineskip}
    \caption{Layer dropping patterns when \sys (calibration + activation-norm tie breaker) is applied to LLaMA-7B on QA benchmarks. Results for the first 32 iterations are shown. At this point, the model has been reduced to one half of its original size with nearly no performance loss, evidenced in Figure~\ref{fig:layer_dropping_teaser}. The numerical value -1 is assigned to discarded layers as accuracy no longer applies.}
    \label{fig:layer_dropping_patterns}
\vspace{-0.5\baselineskip}
\end{figure*}

For each of the downstream task shown in Figure~\ref{fig:layer_dropping_patterns}, there are a few key observations can be made: (1) LLaMA-7B have different layer dropping patterns on different tasks, (2) there are significantly more MLP layers are dropped than the self-attention ones. The first observation provides more empirical evidences for layer-wise specialization while the second for knowledge localization, which argues domain knowledge is stored in MLPs. 

\subsection{Ablation: \sys Robustness on a Similar Domain}
\label{appendix:ablation_similar_domain}

We conduct additional experiments to test \sys‘s robustness. Model fine-tuned with the MedMCQA dataset is validated on another dataset, PubMedQA from the medical knowledge domain. Model fine-tuned using the LexGLUE dataset is tested on the Legalbench benchmark. Results in Table~\ref{tab:ablation_similar} show the model specialized on MedMCQA can perform relatively well on PubMedQA, in comparison with benchmarks from totally different knowledge domains. Same conclusion applies to the model specialized on LexGLUE. 

\begin{table*}[h]
\centering
\caption{
Performance of specialized LLaMA-7B on different but similar benchmarks.The percentage in parenthesis indicates the percentage of total parameters remained in the specialized model.
}
\small
\label{tab:ablation_similar}

\begin{tabular}{c|c|c|c|c}
\toprule
\text{model}  & \text{MedMCQA}   & \text{PubMedQA}   & \text{LexGLUE}  & \text{LegalBench}             \\ 
\midrule
\text{w/o fine-tuning (100\%)}   & 22.4 & 5.2 & 32.1 & 44.9   \\ 
\midrule
\text{MedMCQA specialized (40\%)} & \textbf{47.5} & \textbf{58.9} & 12.4 & 19.8 \\ 
\midrule
\text{LexGLUE specialized (40\%)}& 9.1 & 34.5 & \textbf{39.5} & \textbf{49.4} \\ 
\midrule
\end{tabular}
\end{table*}

\subsection{Ablation: Performance Degradation on Unrelated Domains.} We test specialized models' performance on other domain-specific tasks and it demonstrates a significant performance degradation. Results of each specialized model's performance on other tasks are provided in Table~\ref{tab:ablation_performance}. 

By analyzing the outcomes of both the robustness and generalizability experiments in Section~\ref{appendix:ablation_similar_domain} and Table~\ref{tab:ablation_performance}, it becomes evident that a layer deemed insignificant in one domain is likely to hold similar irrelevance in a closely related domain. This observation provides further empirical support to the concept of layer-wise specialization within LLMs. Intriguingly, the balance between such specialization and a model's overall generalizability presents itself as a captivating avenue for research, meriting exploration in future studies.

\begin{table*}[h]
\centering
\caption{
Performance of specialized LLaMA-7B on other QA benchmarks.The percentage in parenthesis indicates the percentage of total parameters remained in the specialized model.
}
\small
\label{tab:ablation_performance}

\begin{tabular}{cccccc}
\toprule
\text{model}  & \text{PIQA}  &  \text{SciQ}  & \text{MedMCQA}    & \text{LexGLUE}  & \text{FinanceQA}                \\ 
\midrule
\text{w/o fine-tuning (100\%)}  & \textbf{77.4} & 89.7 & 22.4  & 32.1& 33.6  \\ 
\midrule
\text{PIQA specialized (40\%)} & \textbf{77.6} & 81.1 & 14.4 & 17.8  &  18.2  \\ 
\midrule
\text{SciQ specialized (40\%)} & 61.5 & \textbf{91.2} & 18.9& 13.0  &  16.5 \\ 
\midrule
\text{MedMCQA specialized (40\%)} & 54.9 & 78.2 & \textbf{47.5} & 12.4 & 14.8  \\ 
\midrule
\text{LexGLUE specialized (40\%)} & 62.4 & 73.1 & 9.1 & \textbf{39.5}  & 18.3 \\ 
\midrule
\text{FinanceQA specialized (40\%)} & 55.3 & 72.5 & 13.8  & 21.7 & \textbf{41.3} \\ 
\bottomrule
\end{tabular}
\end{table*}

\subsection{Ablation: Different Sparse Update Ratios}
\label{appendix:sparse_update}

As we can see in Table~\ref{tab:complete_method_compare_performance_llama}, results show \sys performs the worst when all layers are updated with a sparse update ratio $r=1$. With a ratio of $r = \frac{1}{4}$, the model can be compressed to a greatest extent with more than 16 decoder layers (out of 32) dropped with nearly no loss in accuracy. 

\begin{table*}[b]
\centering
\caption{
Performance comparison of LLaMA-7B \sys variants on QA benchmarks with different combinations of sparse update techniques (Section~\ref{sec:method:pipeline}) and target selection algorithms (Section~\ref{sec:method:target_selection}). For sparse-FT, the frozen layers are determined by calibration scanning and $r=\frac{1}{4}$.
}
\small
\label{tab:complete_method_compare_performance_llama}
\begin{tabular}{ccccccc}
\toprule
\text{methods}  & \text{PIQA} & \text{SciQ} & \text{MedMCQA}  & \text{LexGLUE} &  \text{FinanceQA}  & \text{Final Mem} $\left( \downarrow \right) $       \\ 
\midrule
\multicolumn{7}{c}{ \text{LLaMA-7B} }\\
\text{w/o fine-tuning}  &  77.4 & 89.7 & 22.4 & 32.1 & 33.6 & 100\% \\
\text{+ Full-FT} & 82.4 & 95.6 & 54.6 & 42.9 & 45.1 & 100\% \\ 
\text{+ Sparse-FT} & 83.1 & 95.4 & 53.7 & 43.3 & 46.9 & 100\% \\ 
\midrule
\multicolumn{7}{c}{LLaMA-7B-\sys $\left(r=1\right)$}  \\ 
+ calibration & 81.1 & 94.2 & 52.1  & 41.8 & 42.5 & $\geq 85\%$ \\ 
+ activation-norm & 78.5 & 93.8 & 51.2 & 40.4 & 41.4 & $\geq 90\%$ \\ 
+ both & 80.5 & 94.0 & 52.7 & 42.0 & 43.3 & $\geq 70\%$ \\
\midrule
\multicolumn{7}{c}{LLaMA-7B-\sys $\left(r=\frac{1}{2}\right)$}  \\ 
+ calibration & 80.2 & 93.6 &  51.5 & 42.3 & 41.9 & $\geq 60 \%$ \\ 
+ activation-norm  & 79.5 & 93.4 & 51.8 & 40.1 & 42.0 & $\geq 85 \%$ \\ 
+ both & 80.6 & 93.9 & 52.5 & 41.4 & 42.2 & $\geq55 \%$ \\
\midrule
\multicolumn{7}{c}{LLaMA-7B-\sys $\left(r=\frac{1}{4}\right)$}  \\ 
+ calibration  & 80.5 & 94.0 & 52.4 & 41.5 & 42.3 & $\geq 55\%$ \\ 
+ activation-norm  & 79.6& 93.5 & 51.5 & 39.8 & 41.7 & $\geq 85\%$ \\ 
+ both & \textbf{81.8} & \textbf{94.2} & \textbf{53.1} & \textbf{42.0} & \textbf{43.6} & $\mathbf{\geq 50\%}$ \\
\midrule
\multicolumn{7}{c}{LLaMA-7B-\sys $\left(r=\frac{1}{8}\right)$}  \\ 
+ calibration & 81.3 & 94.5 & 52.6 & 42.0 & 42.8  & $\geq 70\%$ \\ 
+ activation-norm & 80.4 & 94.0 & 51.9 & 39.6 & 42.3 & $\geq 90 \%$ \\ 
+ both & 81.5 & 94.3 & 52.8 & 41.6 & 43.1 & $\geq 60 \%$ \\
\bottomrule
\end{tabular}
\end{table*}




\begin{table*}[b]
\centering
\caption{
Performance comparison of LLaMA-7B variants when applying AWQ and sparseGPT to \sys on domain-specific tasks. The numerical values are percentage in accuracy. Throughputs are meausred on A100 GPUs and are reported in tokens/s with sequence length 512 and batch size 1.
}
\vspace{0.5\baselineskip}
\small
\label{tab:composite_qa_performance_llama}
\begin{tabular}{cccccc}
\toprule
\text{models} & \text{SciQ} & \text{MedMCQA} &  \text{FinanceQA}  & Throughput & \text{Final Mem $\left( \downarrow \right)$}      \\ 
\midrule
\multicolumn{6}{c}{ \text{LLaMA-7B} }\\
\midrule
\text{w/o training} & 89.7 & 22.4  & 33.6 & 42.3 & 100\% \\ 
\text{+ Full-FT} & 95.6 & 54.6  & 45.1 & 42.3 & 100\% \\ 
\text{+ Sparse-FT}  & 95.4 & 53.7  & 46.9 & 42.3 & 100\% \\ 
\midrule
\text{+ SparseGPT (2:4)}  & 90.1 & 52.3 & 41.6 & 58.9 & $100\% $\\ 
\text{+ AWQ-int4} & 93.0 & 50.7 & 42.1 & 115.3 & $> 25\%$ \\ 
\midrule
\multicolumn{6}{c}{LLaMA-7B-\sys $\left(r=\frac{1}{4}\right)$}  \\ 
\midrule
w/o PT compression  & \textbf{94.2} & \textbf{53.1} & \textbf{43.6} & 103.1 & $\geq 50\%$ \\ 
\text{+ SparseGPT (2:4)}  & 89.1 & 47.8 & 38.9 & 132.0 & $\geq 50\%$ \\ 
\text{+ AWQ-int4} & 91.5 & 49.2 & 40.5 & \textbf{188.7} & $>\textbf{ 12.5 \%}$ \\ 
\bottomrule
\end{tabular}
\vspace{-0.5\baselineskip}
\end{table*}



\begin{table*}[b]
\centering
\caption{
Performance comparison of OPT-1.3B and OPT-6.7B variants on QA benchmarks. The numerical values are percentage in accuracy. \sys here uses the best strategy with sparse update at $r = \frac{1}{4}$, calibration scanning and activation-norm tie breaker. For sparse-FT, the frozen layers are determined by calibration scanning and $r=\frac{1}{4}$. 
}
\vspace{0.5\baselineskip}
\small
\label{tab:qa_performance_opt}
\begin{tabular}{ccccc}
\toprule
\text{models}   & \text{SciQ} & \text{MedMCQA}  & \text{LexGLUE}   & \text{Final Mem $\left( \downarrow \right)$}        \\ 
\midrule
\multicolumn{5}{c}{OPT-1.3B} \\
\midrule
\text{w/o training}  & 84.9  & 12.5 & 18.1 &   100\% \\ 
\text{+ Full-FT} & 91.7  & 46.2 & 24.7 &   100\% \\ 
\text{+ Sparse-FT} & 91.5 & 45.6 & 25.1 &   100\% \\ 
\midrule
\text{+ SparseGPT (2:4)} & 88.3 & 42.1  & 20.0 &  $100\%$\\ 
\text{+ LLM.int8()} & 90.8 & 44.2 & 24.0 &  $> 50\%$ \\
+ \sys (50\%) & 88.4 & 43.9 & 21.3  &   $\geq 50\%$ \\ 
+ \sys (40\%) & 85.0 & 39.0 & 18.5 &  $\geq 40\%$ \\ 
\midrule
\multicolumn{5}{c}{OPT-6.7B} \\
\midrule
\text{w/o training}  & 89.0	 & 14.5 & 22.7 &   100\% \\ 
\text{+ Full-FT} & 95.3	 & 49.8 & 41.0 & 100\% \\ 
\text{+ Sparse-FT} & 94.9 & 48.3 & 40.2 & 100\% \\ 
\midrule
\text{+ SparseGPT (2:4)} & 91.0 & 44.2 & 36.2 &   $100\%$\\ 
\text{+ LLM.int8()} & 95.1 & 46.9 & 39.9 &   $> 50\%$ \\
+ \sys (50\%) & 88.5 & 44.3 & 37.5 &   $\geq 50\%$ \\ 
+ \sys (40\%) & 82.2 &	41.9 & 33.7 & $\geq 40\%$ \\ 
\bottomrule
\end{tabular}
\vspace{-0.5\baselineskip}
\end{table*}

\subsection{\sys Fine-tuning Time Complexity Analysis}
\label{appendix:time-complexity}

For conventional full FT, assume the average time it takes to train a layer for one epoch can be approximated by some parameter $c$. In practice, $c$ is a function of positional index (depth of the layer), parameter size, dataset size, sequence length, operator types, hardware types, and other factors for each layer. Specifically, $c$ differs significantly for the MLP and the attention layers. This difference in forward time can be used to assign MLP and attention layers with different weights in addition to the metrics in Equation~\ref{eq:scan_score} and Equation~\ref{eq:norm_score}. In this analysis, we perform order-of-magnitude approximation, assuming $c$ is given as prior knowledge, and leave the opportunities of dynamically estimating $c$ for future work. 

With this approximation, the time it takes to train $N$ layers for one epoch is $T\left( N \right) = cN$ and $T_{\text{FFT}} = cN \times n$ for $n$ epochs. With one layer dropped at a time, let the total number of layers to be dropped be $n_d$, the time can be approximated as:

\begin{align}
T_{\text{FFT}, \Delta=1}\left( n_d \right) = T(N) + T(N-1) + \dots +T(N-n_d+1)
= c \cdot n_d \cdot \left(N - \dfrac{n_d - 1}{2}\right)
\label{eq:5}
\end{align}

Similarly, we can write down approximated time for dropping two layers at a time, and it amounts to:

\begin{align}
T_{\text{FFT}, \Delta=2}\left( n_d \right) = T(N) + T(N-2) + \dots +T(N-n_d+2)
= c \cdot \dfrac{n_d}{2} \cdot \left(N -  \dfrac{n_d}{2} + 1\right)
\label{eq:6}
\end{align}

\begin{table*}[b]
\centering
\small
\caption{
Evaluation of wall clock time of running \sys and other baselines on 2A100-80G GPUs on the SciQ. We use the best \sys scheme reported in Table~\ref{tab:method_compare_performance_llama}.
}
\begin{tabular}{c|c|c}
\toprule
\text{Methods} &              epochs  &     FT Time (h) \\
\hline 
\multirow{3}{*}{Full FT} &      2    &          4.1         \\
                         &       4     &        8.3         \\
                          &      8     &         16.5        \\
\midrule
\multirow{3}{*}{Sparse FT $\left(r = \frac{1}{4} \right)$} &    2    &            2.5       \\
                           &     4     &          5.0       \\
                           &     8     &          10.0       \\
\midrule
\multirow{3}{*}{\sys} &           2    &          1.8         \\
                           &     4     &          3.4       \\
                           &     8     &          6.5       \\
\bottomrule
\end{tabular}
\label{tab:training_profiling}
\vspace{-0.5\baselineskip}
\end{table*}

On the other hand, if we apply \textbf{sparse FT} introduced in Section~\ref{sec:method:sparse_update}, empirical results show it reduces training time to $\sim 60\%$ at the sparsity ratio of $r=1/4$ as evidenced in profiling results from Table~\ref{tab:training_profiling}. The table shows \sys take around $\sim 40\%$ of the time to train in comparison with full FT, and it's now possible to run the proposed fine-tuning scheme that iteratively compress a model without sacrificing too much training overhead.

\begin{figure*}[h]
\centering
\includegraphics[width=\textwidth]{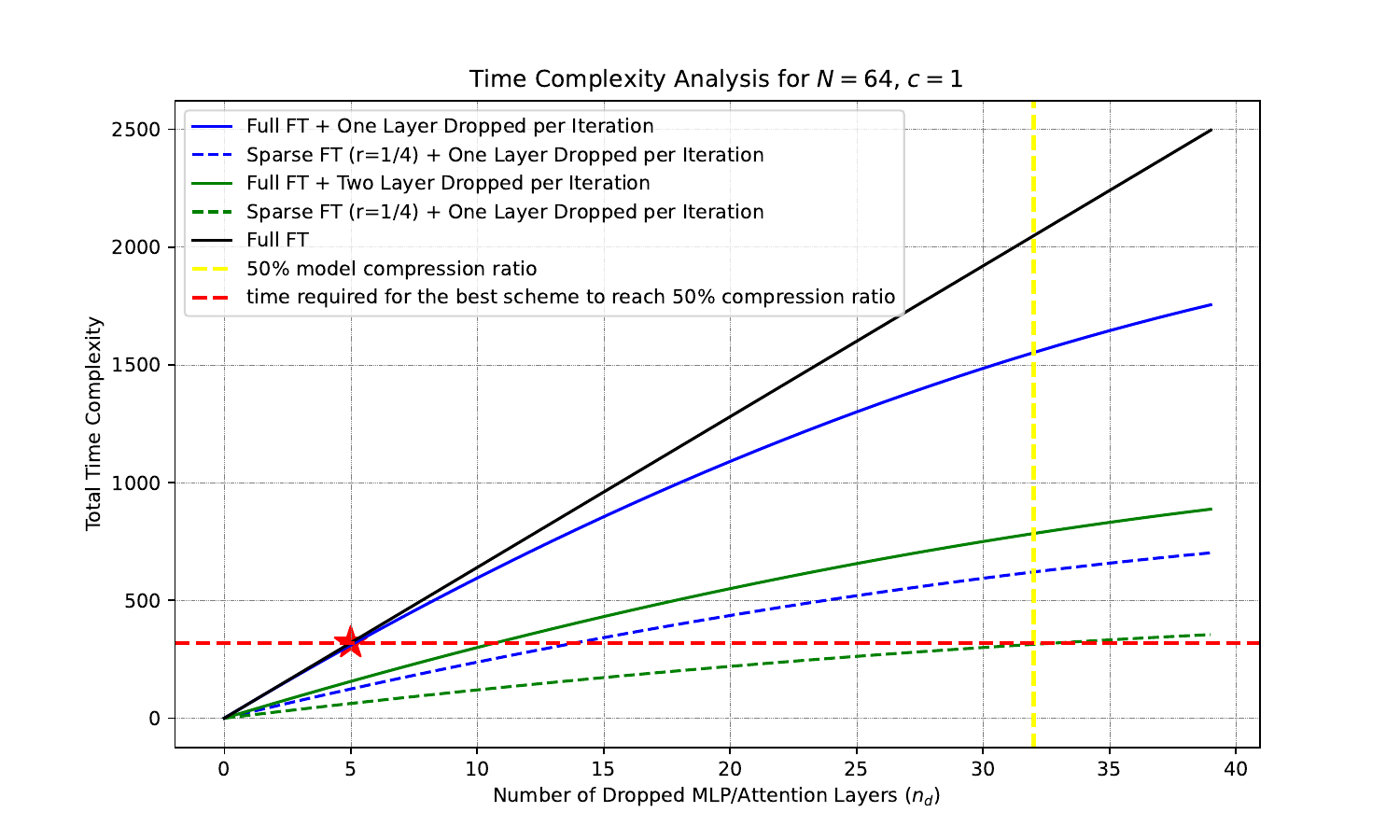} 
\vskip -0.1in
\caption{An illustration of fine-tuning time complexity for different combinations of sparse fine-tuning schemes and number of layers to be dropped versus full fine-tuning. For all curves drawn, we use normalized $c=1$, $N=64$ to simulate the total number of layers in LLaMA-7B. When two layers are dropped per iteration with sparse FT at $r=\frac{1}{4}$, it only requires 5 epochs of fine-tuning to achieve 50\% model compression ratio.
}
\label{fig:time_complexity}
\vskip -0.1in
\end{figure*}

However, it's arguable that the training overhead is still significant as demonstrated in Figure~\ref{fig:time_complexity}. We further introduce \textbf{adaptive layer dropping}, which is to drop more than one layer per epoch. From table~\ref{tab:training_profiling}, we see the conversion factor for training time from full FT to sparse FT is roughly 0.6. Plug the scaling factor into Equation~\ref{eq:5} and Equation~\ref{eq:6}, we can take the estimation $T_{\text{SFT}, \Delta=1}\left( n_d \right) = 0.6 T_{\text{FFT}, \Delta=1}\left( n_d \right)$, and $T_{\text{SFT}, \Delta=2}\left( n_d \right) = 0.6 T_{\text{FFT}, \Delta=2}\left( n_d \right)$. 

Assume $c=1$, we can plot all equation together in Figure~\ref{fig:time_complexity} to compare the training cost of different schemes. If we compare $T_{\text{FFT}, \Delta=2}\left( n_d \right)$ with $T_{\text{FFT}} = cN \times n$, it can be seen that at the 50\% compression ratio, it take roughly the same amount of time to obtain \sys by dropping two layers at a time with sparse update as performing full FT for 5 epochs. This amounts to roughly the same amount of time it takes for standard LLM fine-tuning practice. In addition, our experiments show dropping two layers epoch using the current method is feasible without introducing significant overhead when the number of layer dropped is small. We leave this to our future work for further efficiency improvement. 

\end{document}